\documentclass[letterpaper, 10 pt, conference]{ieeeconf}

\IEEEoverridecommandlockouts
\usepackage{cite}

\usepackage[pagewise]{lineno}
\usepackage{graphicx}
\usepackage{graphicx, subcaption}
\graphicspath{{./figures/}}

\usepackage{comment}
\usepackage{hyperref}
\usepackage{float}
\usepackage[super]{nth}

\title{\LARGE \bf
Leveraging Structure from Motion to Localize Inaccessible Bus Stops
}

\author{Indu Panigrahi$^{1}$, Tom Bu$^{2}$, and Christoph Mertz$^{2}$% <-this % stops a space
\thanks{$^{1}$Indu Panigrahi is with Robotics Institute Summer Scholars
Program at Carnegie Mellon University, Pittsburgh, PA 15213, USA
         and also
with the Department of Computer Science at Princeton
University, NJ 08544, USA
        {\tt\small indup@princeton.edu}}%
\thanks{$^{2}$Tom Bu and Christoph Mertz are with  the Robotics Institute at Carnegie Mellon University, Pittsburgh, PA 15213, USA
        {\tt\small tomb, cmertz@andrew.cmu.edu}}%
}

\begin{document}
% \linenumbers % Uncomment this to enable line numbers in the peer review

\maketitle
\thispagestyle{empty}
\pagestyle{empty}

%%%%%%%%%%%%%%%%%%%%%%%%%%%%%%%%%%%%%%%%%%%%%%%%%%%%%%%%%%%%%%%%%%%%%%%%%%%%%%%%
\begin{abstract}

The detection of hazardous conditions near public transit stations is necessary for ensuring the safety and accessibility of public transit. Smart city infrastructures aim to facilitate this task among many others through the use of computer vision. However, most state-of-the-art computer vision models require thousands of images in order to perform accurate detection, and there exist few images of hazardous conditions as they are generally rare.
% , such as snow coverage on sidewalks near bus stops,

In this paper, we examine the detection of snow-covered sidewalks along bus routes. Previous work has focused on detecting other vehicles in heavy snowfall or simply detecting the presence of snow. However, our application has an added complication of determining if the snow covers areas of importance and can cause falls or other accidents (e.g. snow covering a sidewalk) or simply covers some background area (e.g. snow on a neighboring field). This problem involves localizing the positions of the areas of importance when they are not necessarily visible. 

We introduce a method that utilizes Structure from Motion (SfM) rather than additional annotated data to address this issue. Specifically, our method learns the locations of sidewalks in a given scene by applying a segmentation model and SfM to images from bus cameras during clear weather. Then, we use the learned locations to detect if and where the sidewalks become obscured with snow. After evaluating across various threshold parameters, we identify an optimal range at which our method consistently classifies different categories of sidewalk images correctly. Although we demonstrate an application for snow coverage along bus routes, this method can extend to other hazardous conditions as well. Code for this project is available at \url{https://github.com/ind1010/SfM_for_BusEdge}.

\end{abstract}

\begin{keywords}

Computer Vision for Transportation, Intelligent Transportation Systems, Localization, Segmentation and Categorization
% keywords, choose from \\ https://www.ieee-ras.org/publications/ra-l/keywords

\end{keywords}

%%%%%%%%%%%%%%%%%%%%%%%%%%%%%%%%%%%%%%%%%%%%%%%%%%%%%%%%%%%%%%%%%%%%%%%%%%%%%%%%
\section{INTRODUCTION}
Smart city infrastructures aim to use fields like computer vision to facilitate city management, part of which involves overseeing transportation systems. As transportation systems become more intelligent, an increasing amount of public transit vehicles are equipped with cameras that capture thousands of images of the city per day along with geographic positioning information. City infrastructures can use this immense amount of raw data to monitor the conditions of public transit stations and the surrounding areas.

% Thus far, most work focuses on either identifying the presence of snowy weather or detecting other objects, particularly other vehicles, in snowfall. 
% Furthermore, there is work on an analogous problem of detecting roads in adverse weather conditions, such as snow and fog, in the context of autonomous driving.
% However, the methods developed for this problem are unable to accurately detect roads that are covered by heavy snowfall.
% Snow-covered sidewalks are difficult to recognize; a snow covered sidewalk and a snow-covered grassy area would be indistinguishable to a trained deep learning model. 
% [MOVE]

\begin{figure}[H]
    \centering
    \includegraphics[width=0.55\linewidth]{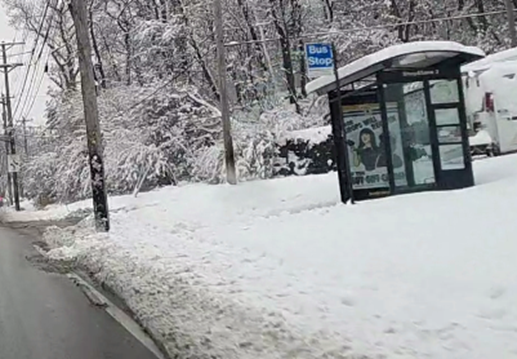}
    \caption{Snow-covered sidewalk leading to a bus stop.}
    \label{fig:snowcoveredbustop}
\end{figure}

Our application focuses on detecting snow-covered sidewalks along bus routes; snow-covered sidewalks are one type of hazardous condition that can limit the safety and accessibility of public buses as pedestrians can lose access to bus stops and/or slip (Fig. \ref{fig:snowcoveredbustop}).
We use images that are captured on-board a public bus as data.
However, instead of annotating this data, we leverage the fact that the bus travels around a set route and apply Structure from Motion and a segmentation model to learn the locations of the sidewalks in clear weather. Then, in future rounds, when the bus encounters snowfall, we compare the detected snow coverage to the learned locations of the sidewalks. If the coverage exceeds a set threshold, we generate an alert, and the bus company can contact the city to clear the sidewalk.

\vspace{-1em}

When evaluating on a few categories of sidewalk images, we identify a set of thresholds at which our method performs well across all categories for this bus route.
Though we demonstrate an application for detecting snow-covered sidewalks, our method can generalize to detecting other conditions such as snow on roads or bike lanes.
% changes in road infrastructures

% Specifically, we use images from one round of the bus route in clear weather and save the segmentation masks of the sidewalks. Furthermore, we use Structure from Motion to produce a reconstruction of the portions of the bus route that contain sidewalks. Then, in future rounds, when the bus could encounter snowy weather. When the bus identifies the presence of a significant amount of snow, we use the SfM point cloud reconstruction to check if the currently-viewed image is present in that point cloud (i.e. whether or not the bus is currently viewing something that is on its route). If the currently-viewed image is present in that point cloud, we take the clear weather sidewalk mask of the image to which it maps the best and check the IoU between the snow mask and the clear sidewalk mask. If the IoU is larger than a given threshold, we classify the image as containing snow on a sidewalk; we test several thresholds of IoU. 
% [MOVE]

% For our evaluation set, we use a 3 class classification. Each image in our evaluation set is labeled as either clear sidewalk, snow-covered sidewalk, or no sidewalk. We calculate the percent accuracy.

% See if this works for one image, then we can extend to an entire route

% Criteria for success:

% Goal is to inform the bus company within an hour. Needs to be approximate. We are ok w false positives

% Bulleted contributions
Our contributions are as follows:
\begin{itemize}
    \item We present a method that combines Structure from Motion with a segmentation model to learn the expected locations of sidewalks and detect whether or not the learned sidewalk locations become covered by snow.
    \item Although we demonstrate by detecting snow-covered sidewalks, our method can easily generalize to other problems.
    \item We collect a small dataset of images depicting sidewalks in clear and snowy weather that we use for evaluation. Additionally, we compile other categories of images that may be relevant for other works.
\end{itemize}

\section{RELATED WORK}
\subsection{\textbf{Existing Municipal Infrastructures}}
Many American cities use the telephone number 311 that allows anyone to report issues for the city to fix, such as snow-covered sidewalks.
However, this process can be inefficient as it is decentralized and relies on the motivation of people.

\subsection{\textbf{BusEdge}}
\begin{figure}[h]
    \centering
    \includegraphics[width=\linewidth]{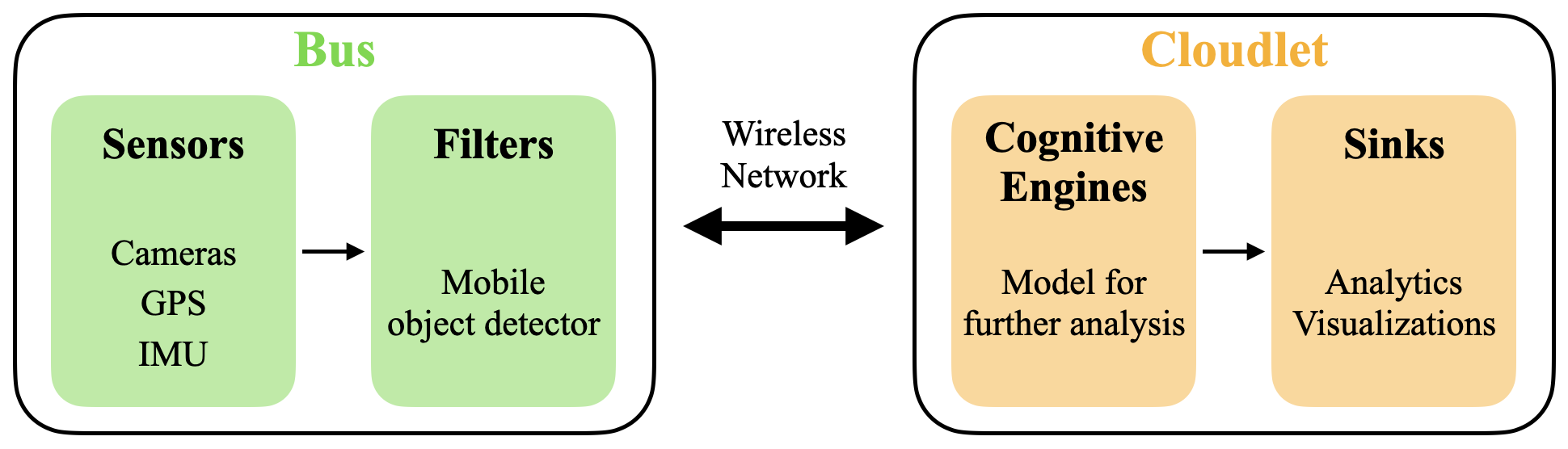}
    \caption{Overview of BusEdge Platform. Figure adapted from Fig. 3.1 in\cite{canbo}.}
    \label{fig:busedge}
\end{figure}

Since buses regularly travel around cities, and many are equipped with cameras, we can facilitate the detection of municipal problems by regularly analyzing bus camera images.
We use a platform called BusEdge\cite{canbo} that captures and packages images with GPS information from the client (bus) and sends the data to the server (cloudlet) to be analyzed (Fig. \ref{fig:busedge}). Intensive on-board analysis can be limited because the bus is equipped with a CPU.

\vspace{-1em}

\subsection{\textbf{Panoptic Segmentation}}
Panoptic segmentation combines semantic and instance segmentation by both categorizing pixels that represent uncountable areas (e.g. snow) and grouping pixels into instances if they belong to countable objects (e.g. cars)\cite{panoptic}.
Although our application involves semantic segmentation categories, we employ a panoptic segmentation model so that our method can be extended more easily for applications where instances are needed.

We apply an off-the-shelf segmentation model called Mask2Former\cite{mask2former}.
This model incorporates a Transformer decoder. 
Transformers have recently become a popular option for computer vision models in terms of accuracy\cite{vit}. 
They are not necessarily more efficient; however, since our application is not significantly time-sensitive (i.e. the bus company can be informed of a snow-covered sidewalk within a few hours rather than within a few seconds), we prioritize accuracy over efficiency.

% Using bc Vision transformers because they should be more accurate. We neglect a bit on efficiency because we are not deploying it immediately. Panoptic transformer can be switched to for future people if they need to detect instances like street lamps instead. Some architectures are accurate

\subsection{\textbf{Snow Detection}}
Most work has focused on detecting the presence of snowfall\cite{weatherconditions1,weatherconditions2,weatherconditions3,weatherconditions4} and localizing the presence of vehicles and other objects in adverse weather conditions such as snow\cite{vehiclesinadverse,vehiclesinrain,vehiclesinsnow}. 
However, in addition to detecting snow, our application has the added complication of localizing the positions of sidewalks that are occluded by snow.

\begin{figure}[H]
    \centering
    \includegraphics[width=0.9\linewidth]{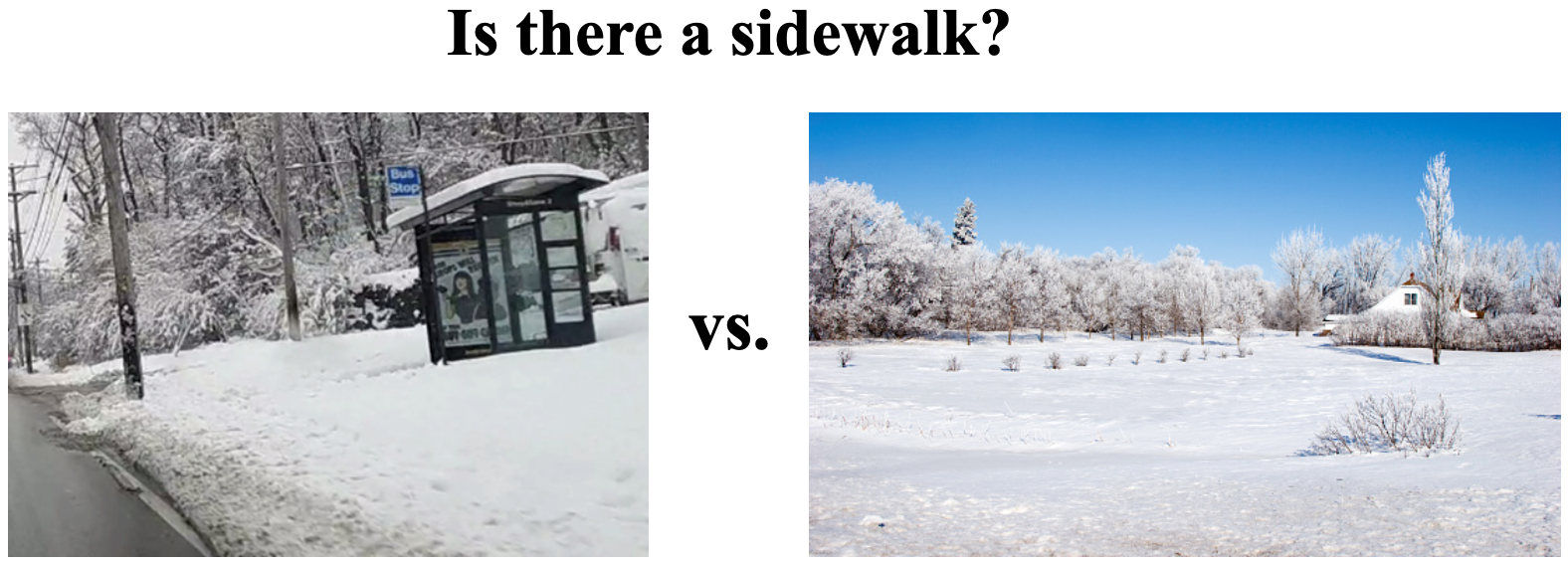}
    \caption{Classifying an image as a snow-covered sidewalk is difficult because the area under the snow is not visible.}
    \label{fig:difficultclassification}
\end{figure}

To our knowledge, there exists no dataset that contains labeled snow-covered sidewalks.
Synthetic images are commonly used to artificially enlarge datasets; however, they are difficult to render realistic-looking\cite{simtorealgap}.
Furthermore, training a deep learning model to classify an image as a ``snow-covered sidewalk" would not be straightforward as any miscellaneous snow-covered area could look identical to a snow-covered sidewalk (Fig. \ref{fig:difficultclassification}).

% No such dataset with snow-covered sidewalk and even if there was it would not be easy to train model

% However, snow-covered sidewalks can look similar to snow-covered grass, so training a model 
% Synthetic images of anomalies could be produced; 

% Difficult to make realistic-looking

% \subsection{\textbf{Localization.}} 

\subsection{\textbf{Image Localization}}
% Faster vs slower

LiDAR is often used to localize the positions of objects surrounding an autonomous vehicle, such as other vehicles\cite{lidar1,lidar2,lidar3,lidar4,lidar5,lidar6}. 
However, LiDAR is expensive, and we already have thousands of images available from bus cameras\cite{lidarcons}.
Furthermore, weather conditions like snow can interfere with LiDAR measurements\cite{lidarcons}.

Some methods have been developed for an analogous problem of localizing roads in adverse weather conditions.
Some applications depend on a previously generated map of the terrain\cite{premademaps1}; we apply a similar idea of generating a preconception of where the sidewalks should be.
A few methods use the geometry of the road, such as the vanishing point of the road and the horizon in the image, to generate an expected target area for where the road could be\cite{geometryroad1, geometryroad2}.
Another method uses self-supervision to generate a pseudo-mask of where the road is expected to be\cite{roadss}.
However, these approaches are more effective in weather conditions under which the road is partially occluded, such as fog or rain.
They are generally unable to localize roads that are fully occluded by snow.
Furthermore, these methods target the autonomous driving domain where they must anticipate completely novel surroundings on any given drive.
On the other hand, we leverage the fact that we work with images from a mostly repetitive bus route.

\begin{figure*}
    \centering
    \includegraphics[width=\textwidth]{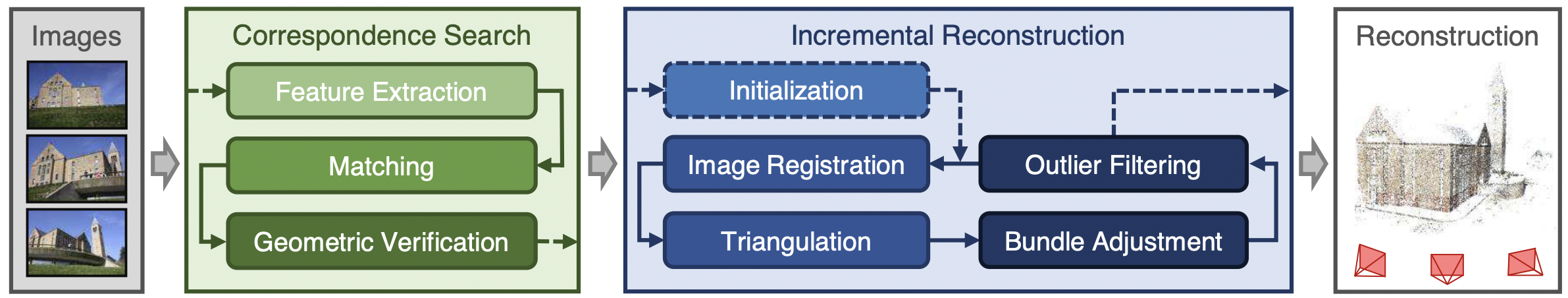}
    \caption{Steps for incremental SfM. Figure from Fig. 2 in\cite{colmap1}.}
    \label{fig:incrementalsfm}
\end{figure*}

Structure from Motion (SfM)\cite{sfm,zisserman} is a classic computer vision algorithm that uses several two-dimensional images taken at different angles of a scene to construct a three-dimensional point cloud representation of the scene. Furthermore, SfM can deduce the pose of the camera for each image and for new images of the same scene\cite{phototourism}. We use a pipeline for SfM called COLMAP\cite{colmap1,colmap2}. More specifically, COLMAP implements incremental SfM which gradually adds images when reconstructing a scene (Fig. \ref{fig:incrementalsfm}); this is as opposed to global SfM\cite{globalsfm}. 

Visual odometry (VO) methods can also localize images \cite{visualodometry} and tend to run faster than COLMAP; however, they are not as accurate. 
% Since we combine information from runs in different weather conditions and since the transition between snow and clear is significant, we need a robust method like COLMAP. 
Furthermore, VO methods that involve deep learning\cite{visualodometrydl} are inherently data-hungry, and our method aims to reduce the amount of annotated data needed. 
Since our application is not significantly time-sensitive, and we need to accurately classify a sidewalk as snow-covered or clear, we require a robust pipeline like COLMAP.
Furthermore, the COLMAP software is well-documented and often referenced as a baseline method by these new methods.
% Changes btwn snow and not snow are significant so need robust. We are combining different runs in different weather. COLMAP is well documented and is baseline method.

\section{METHOD}
For simplicity, we describe our method for one stretch of road that includes a bus stop. 
% For a full bus route, we would associate each stretch with its GPS information. 

% This method generalizes to full bus routes.
% Since GPS is only accurate to about $10$ meters, we end up with GPS to point cloud associations rather than GPS to image frame associations. In this paper, we use a particular segment as an example.

\subsection{\textbf{Data collection}}
\begin{figure}
    \centering
    \begin{subfigure}{\linewidth}
         \includegraphics[width=\linewidth]{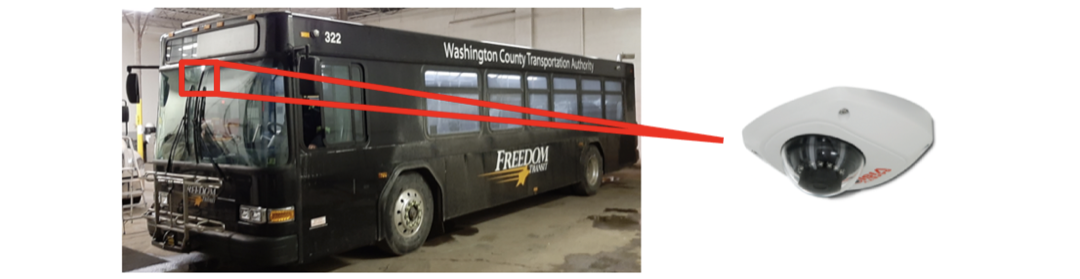}
         \caption{Picture of the bus and camera. See \cite{canbo} for camera specifications.}
         \vspace{0.5em}
         \label{fig:bus}
     \end{subfigure}
     
     \begin{subfigure}{\linewidth}
         \includegraphics[width=\linewidth]{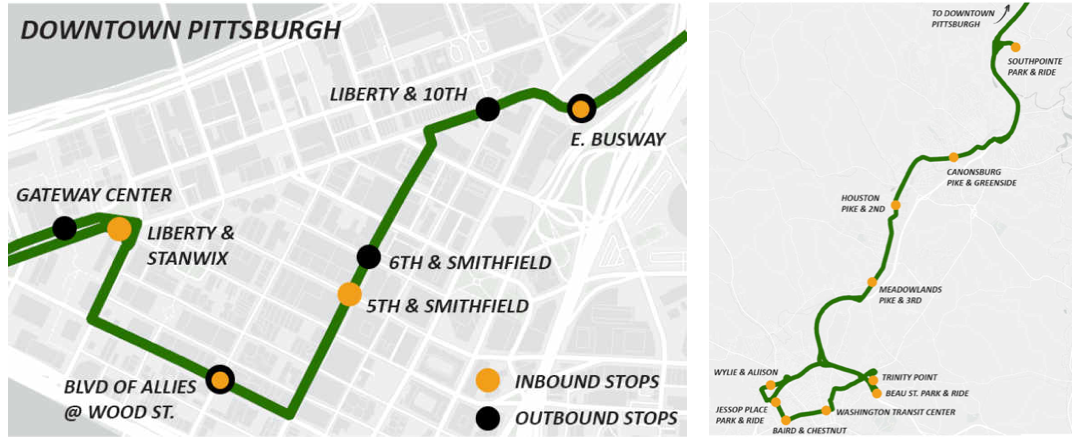}
         \caption{Bus route.}
         \label{fig:route}
     \end{subfigure}
     
    \caption{Information about the bus that is used to collect data.}
    \label{fig:businfo}
\end{figure}

We use images captured from the dash camera on a bus (Fig. \ref{fig:bus}) that travels around Pittsburgh and Washington County, Pennsylvania (Fig. \ref{fig:route}). The camera captures 5 frames per second. Duplicate and blurry images are removed, and the remaining images are sent to a server via the BusEdge Platform\cite{canbo}. The images, along with their GPS and IMU information, are stored as EXIF files in folders of .bag files. We have data beginning from February 2021.

%  which uses the Robot Operating System (ROS)\cite{ros}

% Since snow-covered sidewalks are difficult to find, we first select images from January 7th, 2022 when a snowstorm occurred in the Pittsburgh area.
We choose images corresponding to a stretch of the route with a visible sidewalk in clear weather.
Then, using the GPS information of the selected images, we filter images from other clear-weather days to obtain a few runs of the same sidewalk stretch.

We omit images within the selected GPS range where the bus travels on the opposite side of the road (i.e. returning on the same route) so as to focus on one side of the road.
This omission is not strictly necessary.
Some images have a strong glare from the sunlight and consequently the scenes in these images are extremely dim, so we remove these images.
In the end, we keep three runs of the same stretch of sidewalk in clear weather.
We use a similar process to collect different categories of sidewalk images for evaluation (described in Sec. \ref{section:resultssection}).
% For the clear sidewalk images, we use images from March 29th, 2022 and December 31st, 2021.
% For the snow images, we use images from January 7th, 2022 and inspect the images to find a stretch of sidewalks covered in snow.
%  since that was one time of snow in Pittsburgh using rosbag and BusEdge server-side scripts\cite{canbo} and

\subsection{\textbf{Reconstruction of ground truth sidewalks}}
\begin{figure*}[t]
    \centering
    \begin{subfigure}{0.75\linewidth}
         \includegraphics[width=\linewidth]{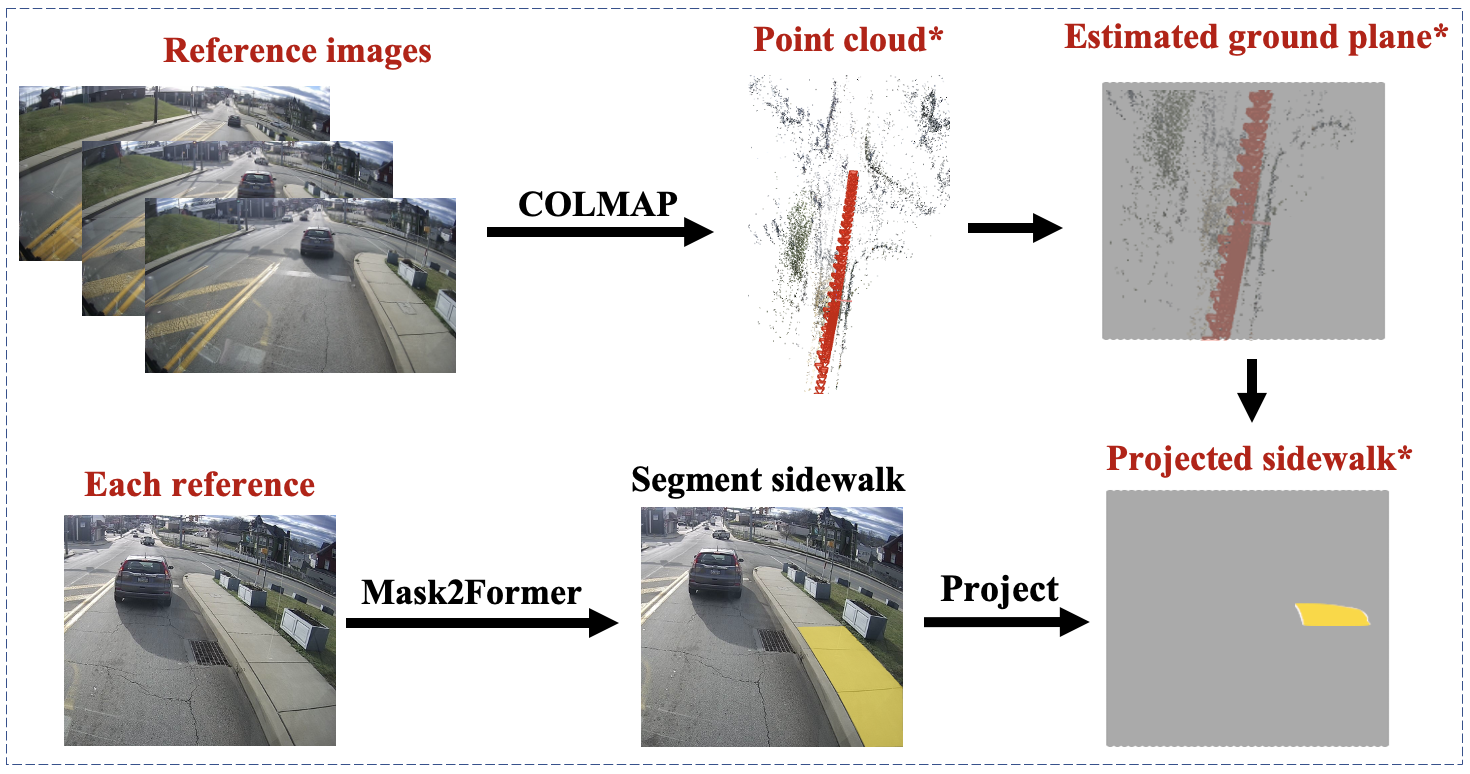}
         \vspace{0.001em}
         \caption{\textbf{Reconstructing the ground truth sidewalks.}}
         \label{fig:firstpart}
     \end{subfigure}
    \begin{subfigure}{0.17\linewidth}
         \includegraphics[width=\linewidth]{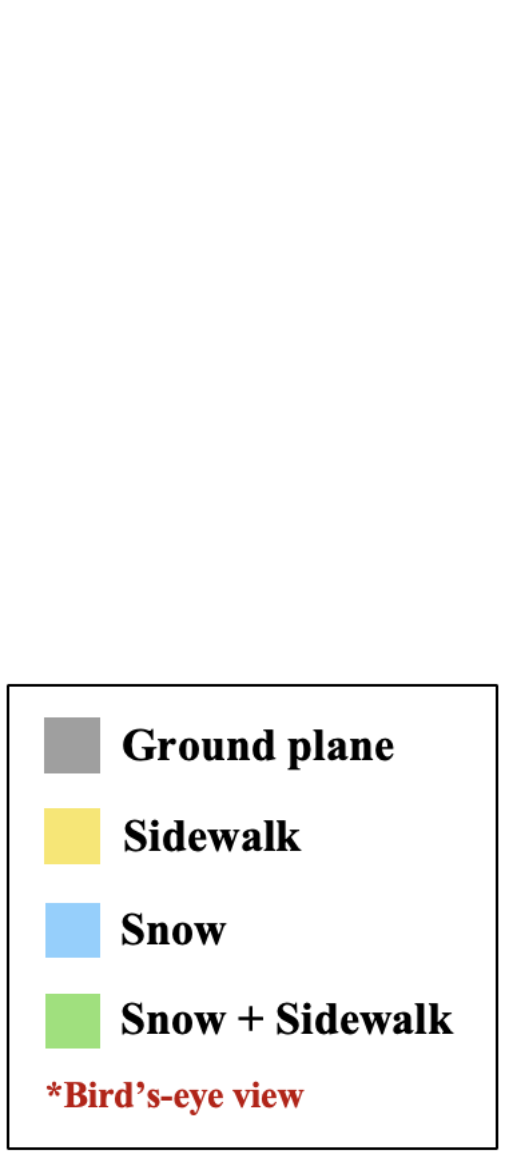}
         \vspace{0.001em}
         \caption{\textbf{Key.}}
         \label{fig:legend}
     \end{subfigure}
     \bigskip
     \bigskip
     
     \begin{subfigure}{\linewidth}
         \includegraphics[width=\linewidth]{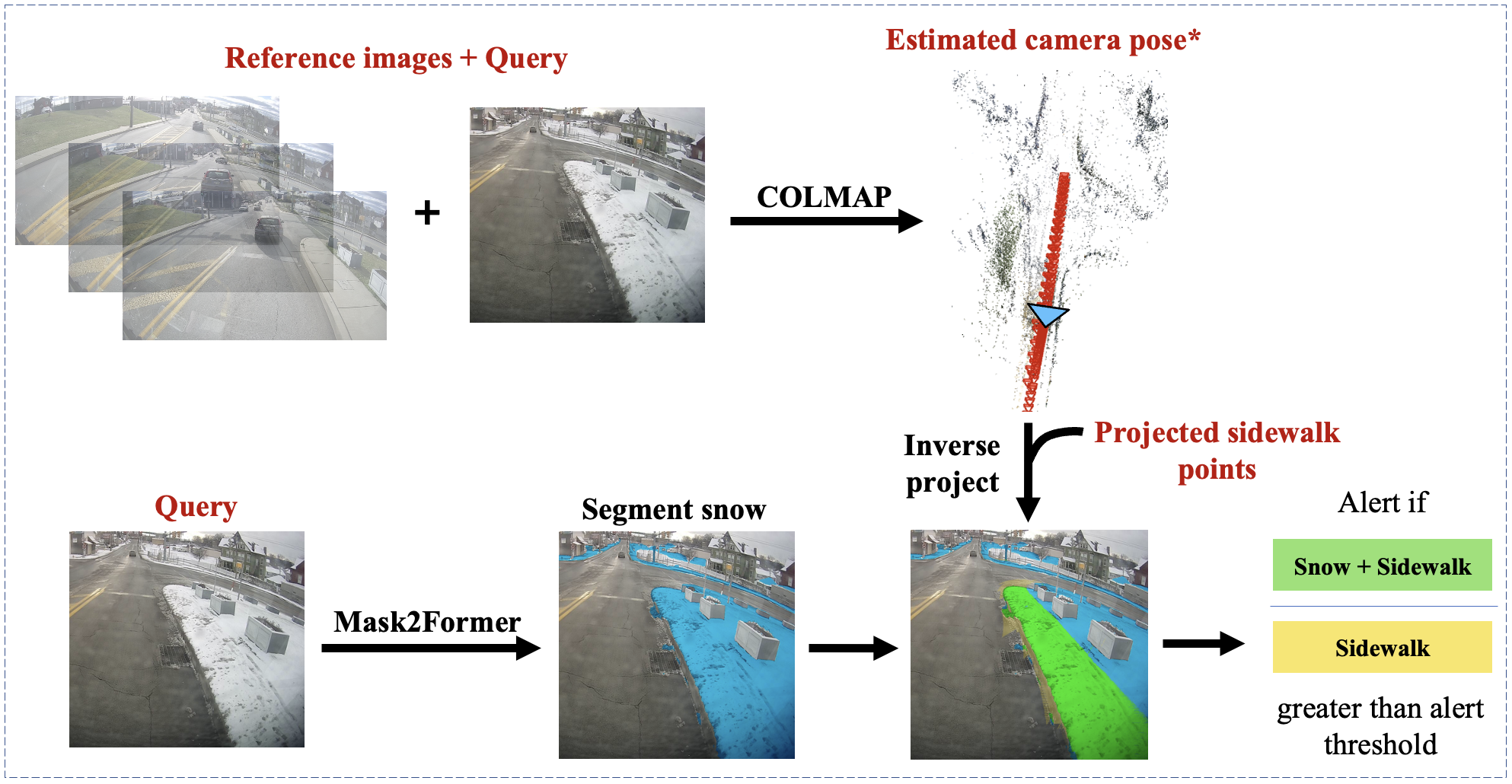}
         \vspace{0.01em}
         \caption{\textbf{Classifying a query image.}}
         \label{fig:secondpart}
     \end{subfigure}
    \vspace{1em}
     
    \caption{We describe this method in terms of one stretch of road that has a bus stop: Our method begins by using clear-weather images of the stretch to render a 3D model of the sidewalks as shown in \textbf{(a)}. This involves running SfM on the reference images to render a point cloud of the scene, estimating the ground plane of the point cloud, and then projecting the portion of the sidewalk mask that is closest to the bus in each reference image onto the estimated plane. When the bus encounters snow coverage, our method compares the snow to the expected sidewalk area as shown in \textbf{(c)}. This process involves estimating the camera pose of the query by re-running COLMAP with the query added to the reference images and using the estimated pose to project the saved sidewalk points into the query image (i.e. the inverse projection of \textbf{(a)}). Finally, if the proportion of the expected sidewalk area that is covered by snow exceeds a set alert threshold, we generate an alert.}
    \label{fig:method}
\end{figure*}

The steps detailed in this section are adapted from an analogous work in the detection of changes in crosswalks\cite{tombu}. For this application, the overall idea is to save the sidewalk locations into a 3D rendering of the scene (Fig. \ref{fig:firstpart}).

\begin{enumerate}
    \item \textbf{Use reference images to render a point cloud of the scene.} \\
    \hspace*{0.2cm} We feed the images of the sidewalk stretch from multiple clear-weather runs (i.e. \textit{reference} images) into COLMAP\cite{colmap1,colmap2} and obtain a point cloud representation of the stretch.
    
    SfM has three main steps\cite{sfmsurvey} (Fig. \ref{fig:incrementalsfm}):
    \begin{enumerate}
        \item \textit{Identify keypoints.} \\
        Keypoints are points in the scene that are somewhat salient and specific to the scene. Objects like monuments and store signs tend to provide robust keypoints.
        \item \textit{Represent the keypoints as vectors.} \\
        COLMAP uses the SIFT descriptor\cite{sift} to extract the features of the keypoints and their local surroundings as vectors \footnote{There also exist other descriptors such as SuperPoint\cite{superpoint} that can be used instead.}.
        \item \textit{Reconstruction.} \\
        First, the pairwise relationships between images are determined by using RANSAC\cite{ransac} and the extracted feature vectors. Then, reconstruction begins with the image pair containing the most inliers. Images are gradually added while solving bundle adjustment.
        % Identify the inliers and the mapping between images
    \end{enumerate}

    \hspace*{0.2cm} We run COLMAP on images from multiple runs of the same stretch because one run tends to produce too sparse a point cloud. 
    Since the bus camera is unlikely to be in the exact same orientation between runs (e.g. not always centered in the lane), we are effectively guaranteed reasonable stereo pairs which improves the 3D reconstruction.
    
    \item \textbf{Estimate the ground plane of the scene.} \\
    \hspace*{0.2cm} First, we segment the road in each reference image by applying the Mask2Former\cite{mask2former} model. 
    We select the panoptic segmentation model with a Swin-L (IN21k) backbone that is pre-trained on Mapillary Vistas. Mapillary Vistas\cite{mapillaryvistas} is a dataset that contains street-level images, and its panoptic segmentation categories include snow, sidewalks, and roads. 
    
    \hspace*{0.2cm} Then, we use the pixel-to-3D point correspondences provided by COLMAP to identify the points in the point cloud that correspond to the road pixels. Next, we use RANSAC to fit a plane to the identified points; this plane is the estimated ground plane. Finally, we re-orient all the points such that the z-axis of the point cloud aligns with the normal of the ground plane.

    \item \textbf{Segment the sidewalk in each reference image.} \\
    \hspace*{0.2cm} For this step, we obtain masks of the sidewalks in the reference images by again applying the Mask2Former model\cite{mask2former}. Using the assumption that the bus is driving on the right side of the road as is the convention in the United States, we omit identified sidewalk pixels that are to the left of and/or above the midpoint of the image. This helps restrict the view of the bus to the sidewalk closest to it. The driving assumption would need to be adjusted in locations where the driving conventions differ.
    
    \item \textbf{Use the estimated ground plane to project the sidewalk masks into the point cloud and save the projected points.} \\
    \hspace*{0.2cm} In order to save the sidewalk locations into the point cloud, we use the road to determine the homography from the image to the point cloud. 
    We can assume that the road is a flat reference area and that the slight lift of the sidewalk does not contribute much error as seen in our example qualitative results (Fig. \ref{fig:qualitative}).
    
    \hspace*{0.2cm} Let us consider one reference image. First, we find the homography from the road pixels in the image to the corresponding 3D points that lie on the estimated ground plane. This is effectively the homography from the image to the estimated ground plane. Next, we use the homography matrix to project each pixel in the sidewalk mask onto the estimated ground plane. Lastly, we save coordinates of the projected sidewalk points. We repeat this process for each reference image, and the combined points form a 3D model of the expected sidewalk locations.
\end{enumerate}

% Next, we save the sidewalk locations in terms of their projected locations on the point cloud.
% Specifically, we obtain a segmentation mask of the road  in each image that was used for reconstruction with the Mask2Former and identify which points in the point cloud correspond to the mask.
% This is done using the image and 3D point IDs provided by COLMAP.
% Using the identified points, we estimate the ground plane of the road.
% Finally, we obtain the segmentation mask of the sidewalks and project the mask onto the estimated ground plane.
% We save the projected coordinates as the expected sidewalk locations.

% where the sidewalks should be.
% We run COLMAP on the combined images from both runs and

% When the segmentation model identifies a significant amount of sidewalk  the images are fed into COLMAP and the sidewalk masks for the images that are used in the reconstruction are saved.
% (maybe change to if the model identifies any sidewalk because knowing if it's a significant amount might involve determining if the sidewalk is relatively salient),

\subsection{\textbf{Classification of query image}}
% Localize snowfall image in point cloud and compare to expected sidewalk locations
In our application, \textit{query} images are images from future runs of the bus when there could be snowfall.
This part of the method involves classifying a query image as Clear or Snow-covered (Fig. ~\ref{fig:secondpart}). 

\begin{enumerate}
    \item \textbf{Check if the query image belongs to the scene.} \\
    For a given query image, we use GPS information to check if the query belongs to the point cloud.
    If the query is within the GPS range for the scene, we proceed.
    
    \item \textbf{Identify the snow coverage in the query, if any is present.} \\
    We use the Mask2Former model\cite{mask2former} to segment the snow in the query and proceed if snow is present.
    
    \item \textbf{Estimate the camera pose of the query.} \\
    If the query does belong to the scene, we add the query to the collection of reference images and re-run COLMAP to obtain an estimated camera pose for the query. Sometimes, COLMAP needs to be re-run more than once to obtain an accurate pose for the query. This reconstruction does not take as long as the initial point cloud rendering because the query is simply added to the existing reconstruction.
    
    \item \textbf{Compare the snow coverage to the expected sidewalk area.} \\
    Using the estimated camera pose and ground plane, we project the saved sidewalk points from the point cloud into the image.
    Finally, we calculate the proportion of the projected sidewalk that overlaps with the snow.
    
    \item \textbf{Generate an alert if the snow significantly covers the expected sidewalk area.} \\
    If the coverage is greater than a set threshold, we generate an alert. See Sec. \ref{section:resultssection} for details about selecting an alert threshold.
\end{enumerate}

% We perform these projections with the rendered point cloud because comparing the masks directly would likely lead to a large amount of error.
% The bus is not always in exactly the same location between runs and consequently the masks would shift.

%  computing the SIFT descriptor for the new image and checking how many inliers it has. If the new image has more than a certain percentage of inliers with any of the images in the point cloud closest to its GPS location, we segment the snow in the new image and check the IoU between the snow mask and the saved sidewalk mask for the point cloud image that it best maps to. If the IoU is greater than a given threshold, we label the image as containing a snow-covered sidewalk. We test different thresholds of inlier percentages and as well as different IoU thresholds.

% Since this project deals with one point cloud, and we explicitly select a new image that does correspond to the same area, we do not test out the choosing the point cloud with the closest GPS location aspect of this.

\section{RESULTS}
\label{section:resultssection}
We evaluate by first reconstructing the ground truth sidewalks for two stretches that include bus stops. These stretches were chosen based on where there were images from the categories described below.
We select and evenly split $66$ test images taken during daylight hours into three categories:

\begin{figure} [H]
    \centering
    \begin{subfigure}{0.3\linewidth}
         \includegraphics[width=\linewidth]{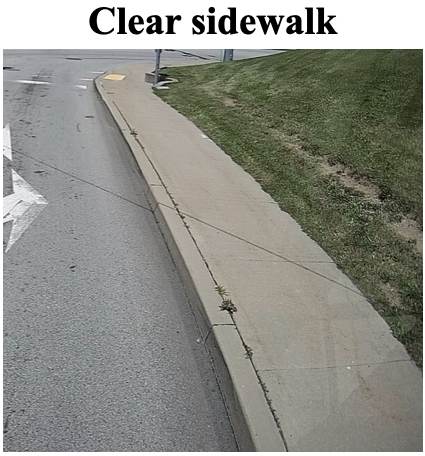}
         \caption{}
         \label{fig:clearex}
     \end{subfigure}
     \begin{subfigure}{0.3\linewidth}
         \includegraphics[width=\linewidth]{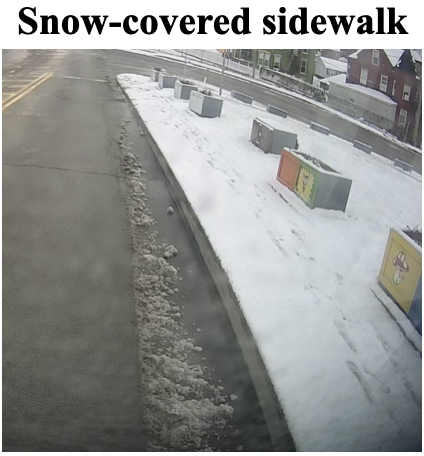}
         \caption{}
         \label{fig:snowex}
     \end{subfigure}
     \begin{subfigure}{0.3\linewidth}
         \includegraphics[width=\linewidth]{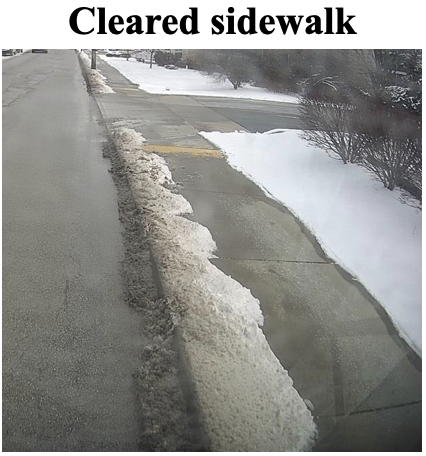}
         \caption{}
         \label{fig:clearedex}
     \end{subfigure}
     
    \caption{Categories of test images. \textbf{(a)} is clear, \textbf{(b)} is snow-covered, and \textbf{(c)} is cleared.}
    \label{fig:examples}
\end{figure}

\begin{enumerate}
    \item \textbf{Clear} \\
    This category includes images in clear weather from February 2022 when the sidewalk in the chosen stretch is clear (Fig. \ref{fig:clearex}). These images should be classified as Clear and are distinct from the reference images originally used to render the point cloud.
    
    \item \textbf{Snow-covered} \\
    This category includes images in which the sidewalks are obscured by snow (Fig. \ref{fig:snowex}). These images should be classified as Snow-covered and were taken on January \nth{7}, 2022 when a snowstorm occurred in the Pittsburgh area.
    
    \item \textbf{Cleared} \\
    This category includes images in which the sidewalks are surrounded by but not covered with snow (i.e. the sidewalks have been cleared) (Fig. \ref{fig:clearedex}). These images are important because snow is present in the image but does not obstruct the sidewalk. These images should be classified as Clear; we do not need to distinguish between clear and cleared sidewalks. Like the snow-covered images, these images were taken on January \nth{7}, 2022 though in a different stretch.
\end{enumerate}

\begin{figure}
    \centering
    \begin{subfigure}{\linewidth}
         \includegraphics[width=\linewidth]{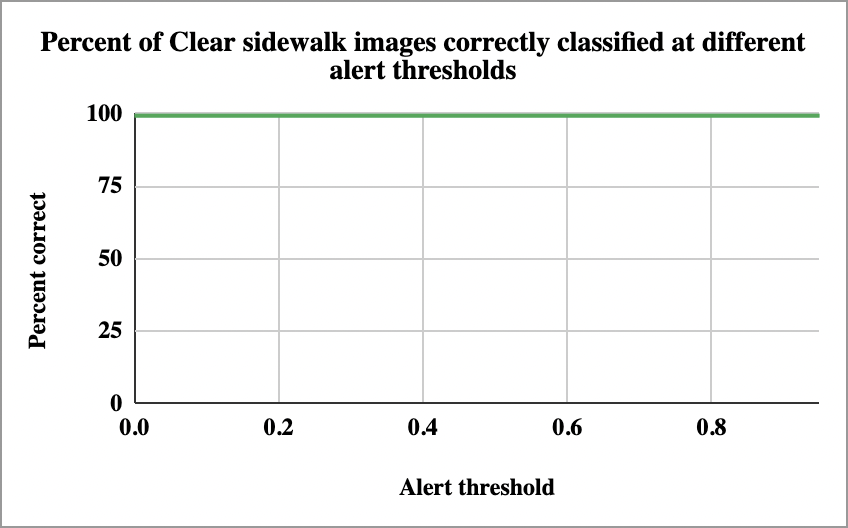}
         \caption{\textbf{Results for clear sidewalks.} Our method trivially performs well at all thresholds because there is no snow in images from this category.}
         \vspace{0.5em}
         \label{fig:clearresults}
     \end{subfigure}
     
     \begin{subfigure}{\linewidth}
         \includegraphics[width=\linewidth]{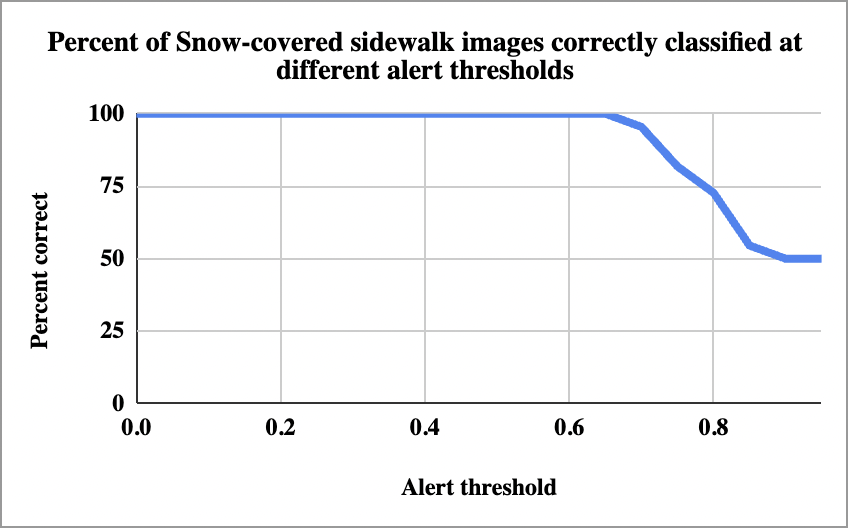}
         \caption{\textbf{Results for snow-covered sidewalks.} Our method performs well until around a threshold of $0.70$ when the percent of images classified correctly begins decreasing more rapidly.}
         \vspace{0.5em}
         \label{fig:snowresults}
     \end{subfigure}
     
     \begin{subfigure}{\linewidth}
         \includegraphics[width=\linewidth]{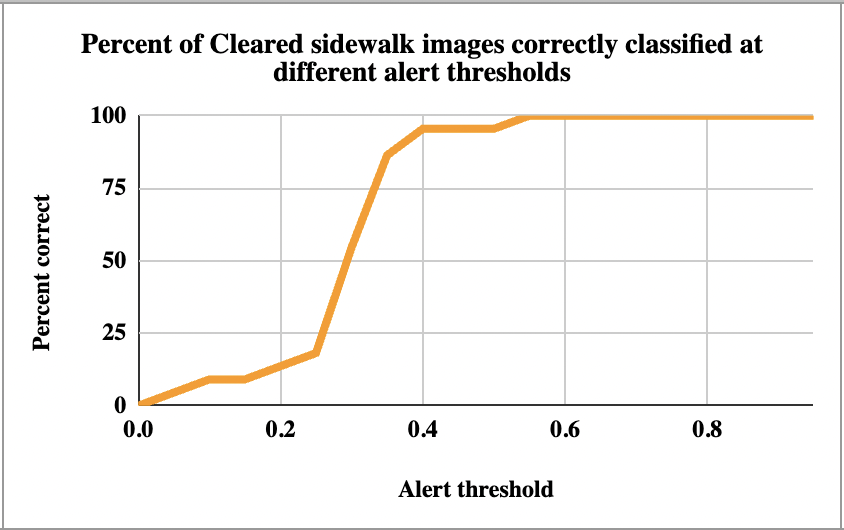}
         \caption{\textbf{Results for cleared sidewalks.} Our method performs well after a threshold of $0.40$ when the percent of images classified correctly begins to stabilize.}
         \label{fig:clearedresults}
     \end{subfigure}
     
    \caption{These graphs depict the percent of images correctly classified across alert thresholds from $0$ to $0.95$ incremented by $0.05$. \textbf{(a)} shows results for clear sidewalks, \textbf{(b)} shows results for snow-covered sidewalks, and \textbf{(c)} shows results for cleared sidewalks. Alert thresholds ranging from $0.58$ to $0.62$ produce an optimal performance across all three test categories.}
    \label{fig:results}
\end{figure}

For each category, we obtain the percent of images that are correctly classified as either Snow-covered or Clear across incremented alert thresholds.
For clear images, our method trivially performs well across all thresholds because there is no snow present in any of the images (Fig. \ref{fig:clearresults}).
For snow-covered sidewalks, our method performs well until a threshold of around $0.7$ (Fig. \ref{fig:snowresults}). This trend is reasonable because a snow-covered sidewalk will have a high, but not necessarily perfect, overlap with the snow in the image.
Finally, for cleared sidewalks, our method performs better as the threshold increases (Fig. \ref{fig:clearedresults}). This trend is reasonable because a stricter (i.e. higher) threshold will classify more images as Clear.

\begin{figure}
    \centering
    \begin{subfigure}{\linewidth}
         \includegraphics[width=\linewidth]{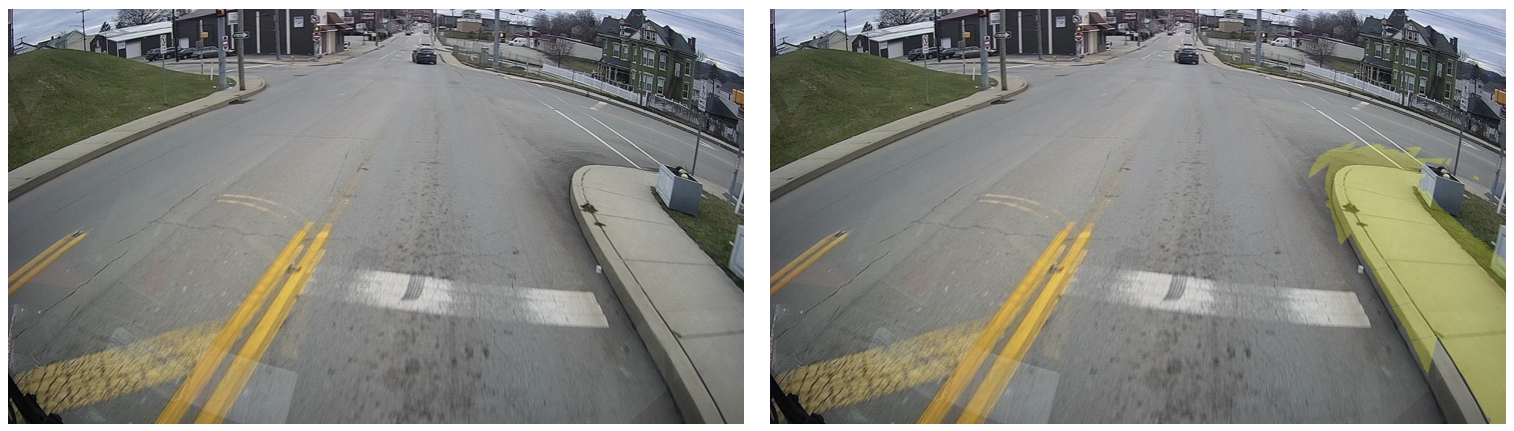}
         \caption{This is an example of a clear query that is correctly classified as Clear (coverage = $0$). No snow is identified in the image, and the projected sidewalk (shown in yellow) aligns fairly well with the actual sidewalk.}
         \vspace{0.5em}
         \label{fig:clearqresults}
     \end{subfigure}
     
     \begin{subfigure}{\linewidth}
         \includegraphics[width=\linewidth]{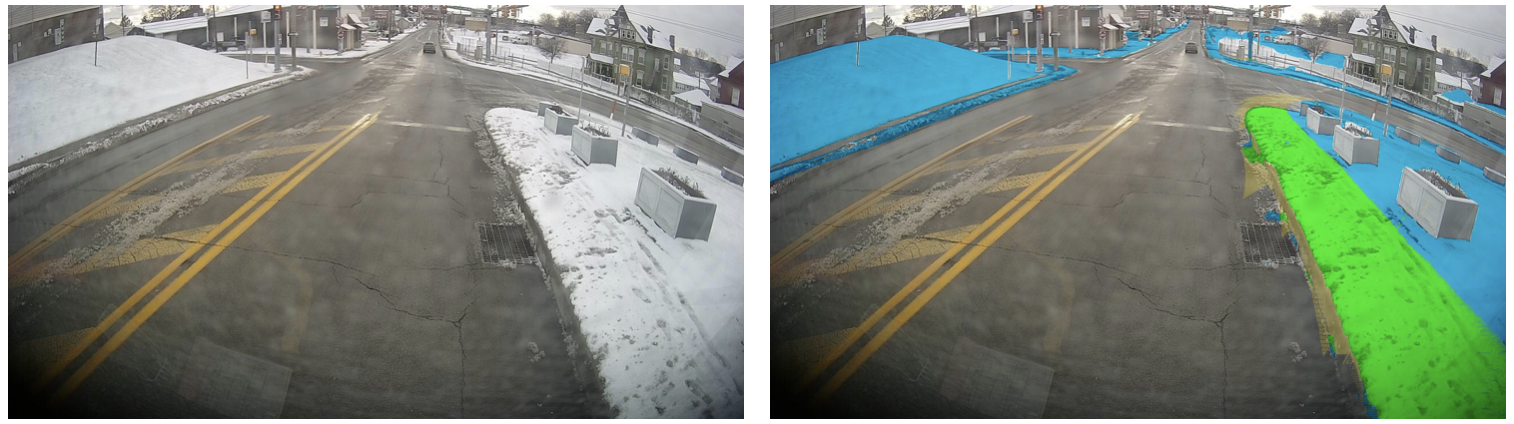}
         \caption{This is an example of a snow-covered query that is correctly classified as Snow-covered (coverage = $0.91$). The projected sidewalk (mostly colored in green due to overlap with snow) aligns almost perfectly with the actual sidewalk which is not visible.}
         \vspace{0.5em}
         \label{fig:snowqresults}
     \end{subfigure}
     
     \begin{subfigure}{\linewidth}
         \includegraphics[width=\linewidth]{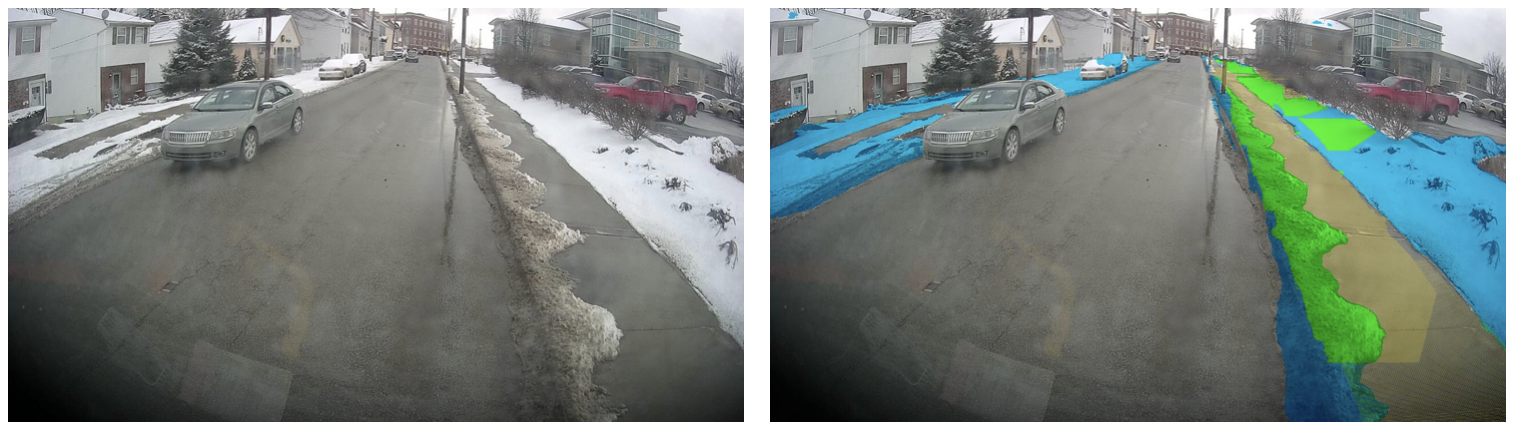}
         \caption{This is an example of a cleared query that is correctly classified as Clear (coverage = $0.36$). The projected sidewalk aligns, though not perfectly, with the sidewalk in the image. The misalignment present towards the top left can occur due to an inaccurate estimated camera pose for the query or for some of the reference images when reconstructing the ground truth sidewalks. In this case, the latter probably occurred as there are full patches of projected sidewalk off to the side.}
         \label{fig:clearedqresults}
     \end{subfigure}
     
    \caption{These images depict an example of a qualitative result for each of the three test categories at an alert threshold of $0.60$. The left panels depict a query image from each category. The right panels display snow in blue, the saved sidewalk projected into the image in yellow, and overlap between snow and the projected sidewalk in green.}
    \label{fig:qualitative}
\end{figure}

Since we need a threshold that will perform well across all categories, we identify $0.58$ to $0.62$ as a good range of thresholds. We include some example qualitative results from each test category at an alert threshold of $0.60$ (Fig. \ref{fig:qualitative}). The saved sidewalk points that are projected into the query image generally align well with the real sidewalk in the query. The few misalignments that we observe (Fig. \ref{fig:clearedqresults}) are most likely due to an inaccurate estimated camera pose for the query and/or for some of the reference images when projecting the ground truth sidewalk masks into the point cloud.

\section{CONCLUSIONS}
In this paper, we present and demonstrate a less data-intensive method for detecting snow-covered sidewalks along bus routes. Our method leverages Structure from Motion to learn the expected locations of sidewalks during clear weather and then uses the learned locations to determine if the sidewalks become covered with snow.
By evaluating on different categories of sidewalks, we identify a range of thresholds across which this method performs well for our bus route.

For our particular application, an immediate extension of this method is to incorporate GPS information to form a full route of point clouds.
However, our method can also extend to other hazardous conditions such as snow-covered bike lanes or roads.

One limitation of this method is its dependence on keypoints.
The effectiveness of the SIFT descriptors in the SfM process depends on the presence of robust and unique keypoints. 
This, in turn, can affect the estimated camera poses for each image.

In urban scenes, there exist many buildings, signs, and sometimes monuments that provide such keypoints. 
However, there may not exist many salient keypoints in rural areas
Likewise, in night settings, keypoints can be less visible.
In these cases, it would be interesting to experiment with adding GPS information for feature matching in SfM.

\section*{ACKNOWLEDGMENT}

This research was supported by the National Science Foundation under Award No. 2038612. Data and background software were provided by projects sponsored by Carnegie Mellon University's Mobility21 National University Transportation Center, which is sponsored by the United States Department of Transportation. We would also like to thank Dr. John M. Dolan, Rachel Burcin, and the RISS program and sponsors for further supporting this project.

% The preferred spelling of the word `acknowledgment' in America is without an `e' after the `g'. Avoid the stilted expression, `One of us (R. B. G.) thanks . . .'  Instead, try `R. B. G. thanks'. Put sponsor acknowledgments in the unnumbered footnote on the first page.

%%%%%%%%%%%%%%%%%%%%%%%%%%%%%%%%%%%%%%%%%%%%%%%%%%%%%%%%%%%%%%%%%%%%%%%%%%%%%%%%

\nocite{*}  % Without this, cite articles in text using \cite{...}
\bibliographystyle{IEEEtran}
\bibliography{./IEEEfull,refs}

\end{document}